%% file: main.tex
\pdfoutput=1
\documentclass[11pt]{article}

\usepackage[]{EMNLP2022}

\usepackage{times}
\usepackage{latexsym}

\usepackage[T1]{fontenc}
\usepackage[utf8]{inputenc}

\usepackage{microtype}

\usepackage{inconsolata}
\usepackage{color}
\usepackage{array}
\newcolumntype{P}[1]{>{\centering\arraybackslash}p{#1}}

\newcommand{\hlc}[2][yellow]{{%
		\colorlet{foo}{#1}%
		\sethlcolor{foo}\hl{#2}}%
}

\usepackage{trackchanges}
\addeditor{YJ}
\addeditor{HC}
\addeditor{WD}

\input{mydef}

\title{Explaining Predictive Uncertainty by Looking Back at Model Explanations}

\author{Hanjie Chen, Wanyu Du, Yangfeng Ji \\
         Department of Computer Science \\ University of Virginia \\  
         Charlottesville, VA 22904\\
        \texttt{\{hc9mx, wd5jq, yangfeng\}@virginia.edu} \\
         }

\begin{document}
\maketitle

\begin{abstract}
  \input{abstract}
\end{abstract}

\input{intro}

\input{relate}

\input{method}

\input{exp}

\input{con}

\section*{Limitations}
One limitation is that we adopted two perturbation-based explanation methods, Leave-one-out and Sampling Shapley, identifying word-level features. 
Utilizing high-level explanation methods (e.g., hierarchical explanations \citep{chen-etal-2020-generating-hierarchical}) may capture more semantic information that explains model predictive uncertainty. 
Another limitation is that we identified uncertain words by removing them and observing whether model prediction confidence increases. 
An alternative way is replacing those words with their synonyms, hence maintaining the original semantic meaning. 
But this may lead to adversarial examples, which we leave to future work. 

\section*{Ethics Statement}
Regarding ethical concerns, this work utilized a sensitive dataset (Wikipedia Toxicity Corpus \citep{wulczyn2017ex}) which contains toxic comments. 
Before conducting human evaluation on this sensitive dataset, we had reported potential participant risks to Institutional Review Board (IRB) and gotten approval of continuing this research. 
We will provide the link to IRB approval with the publication of this paper. 
The other two datasets, IMDB and Senator Tweets, do not have higher risks than those encountered in daily life and daily online activities. 
For all human evaluation experiments, we did not collect any personal information (e.g. demographic and identity characteristics) of participants. 

% \section*{Acknowledgements}

\bibliography{ref}
\bibliographystyle{acl_natbib}

\clearpage
\newpage
\appendix
\input{appendix}

\end{document}

%% file: mydef.tex
\usepackage{amsmath,amsfonts,amssymb,bm}
\usepackage{pifont} % http://ctan.org/pkg/pifont
\usepackage{graphicx,subfigure,epsfig,fancybox} % figures & graph
\usepackage{float}
\usepackage{color} % color
\usepackage{multirow}
\usepackage{natbib}

\usepackage{tikz}
\usetikzlibrary{shapes.geometric, positioning, calc}
\usetikzlibrary{arrows,shapes,calc}
\usetikzlibrary{trees,positioning,mindmap,shadows,fit}
\usetikzlibrary{decorations.pathreplacing}
\usetikzlibrary{tikzmark} % for annotating tables
\usetikzlibrary{intersections} % for pyramid

\newcommand{\revised}[1]{\textcolor{blue}{}}
\renewcommand{\vec}[1]{\boldsymbol{#1}}

\usepackage{adjustbox}
\usepackage{array}
\usepackage{booktabs}

\newcolumntype{R}[2]{%
    >{\adjustbox{angle=#1,lap=\width-(#2)}\bgroup}%
    l%
    <{\egroup}%
}

\usepackage{setspace}

%% file: abstract.tex
Predictive uncertainty estimation of pre-trained language models is an important measure of how likely people can trust their predictions. 
However, little is known about what makes a model prediction uncertain. 
Explaining predictive uncertainty is an important complement to explaining prediction labels in helping users understand model decision making and gaining their trust on model predictions, while has been largely ignored in prior works.
In this work, we propose to explain the predictive uncertainty of pre-trained language models by extracting uncertain words from existing model explanations. 
We find the uncertain words are those identified as making negative contributions to prediction labels, while actually explaining the predictive uncertainty. 
Experiments show that uncertainty explanations are indispensable to explaining models and helping humans understand model prediction behavior.

%% file: intro.tex
\section{Introduction}
\label{sec:intro}
Pre-trained language models \citep[e.g., BERT;][]{devlin-etal-2019-bert} have been indispensable to natural language processing (NLP) due to their remarkable performance \citep{liu2019roberta, yang2019xlnet, gururangan-etal-2020-dont, brown2020language}. 
Predictive uncertainty estimation of pre-trained language models is an important measure of how likely people can trust their predictions \citep{desai-durrett-2020-calibration, xu-etal-2020-understanding-neural}. 

A typical way of measuring predictive uncertainty is to calibrate model outputs with the true correctness likelihood \citep{guo2017calibration, kong-etal-2020-calibrated, zhao2021calibrate}, so that the output probabilities well represent the confidence of model predictions. 
In this case, higher prediction confidence indicates lower uncertainty \citep{xu-etal-2020-understanding-neural, jiang2021can}. 

\begin{figure}[t]
  \centering
  \includegraphics[width=0.48\textwidth]{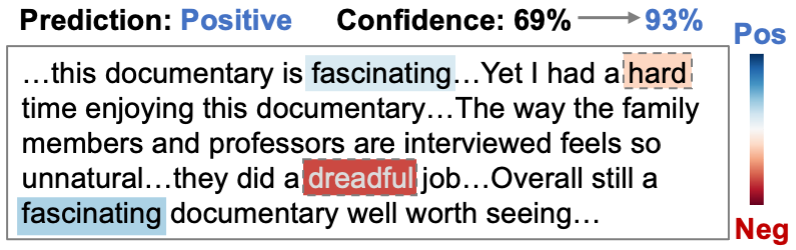}
  \caption{\label{fig:illustrations} An illustration of model explanation for sentiment classification, where the model makes the correct prediction (\textsc{positive}) with a relatively low confidence $69\%$. The top and bottom salient words with respect to the predicted label are highlighted in blue and red colors respectively, indicating different sentiment polarities. Darker color implies larger attribution. Removing the two bottom salient words in dashed boxes can improve the model prediction confidence to $93\%$.}
\end{figure}

However, little is known about what makes a model prediction uncertain. 
Explaining predictive uncertainty is important to understanding model prediction behavior and complementary to explaining prediction labels for gaining users' trust, while has been largely ignored \citep{antoran2020getting}. 
Most works on model explanations focus on explaining a model from the post-hoc manner by identifying important features in inputs that contribute to model predicted labels \citep{ribeiro2016should, lundberg2017unified, sundararajan2017axiomatic, chen-etal-2020-generating-hierarchical, chen-etal-2021-explaining}. 
\autoref{fig:illustrations} shows an example of model explanation for sentiment classification, where the model makes the correct prediction (\textsc{positive}) with a relatively low confidence $69\%$.
The top two salient words highlighted in blue color explain the predicted label. 
However, users may still wonder what compromises the prediction confidence?

This work is the first to explain model predictive uncertainty in NLP. 
Specifically, this work is based on a simple observation that bottom salient words in model explanations (e.g., \texttt{dreadful} and \texttt{hard} in \autoref{fig:illustrations}) identified as making negative contributions to predicted labels actually explain model predictive uncertainty. 
The two bottom salient words in \autoref{fig:illustrations} indicate the opposite sentiment (\textsc{negative}) to the model predicted label. 
Removing them can improve the model prediction confidence from $69\%$ to $93\%$. 
We argue that both top and bottom salient words are indispensable to explaining model predictions. 
We name top salient words as \emph{important words}, explaining model predicted labels; and bottom salient words as \emph{uncertain words}, explaining model predictive uncertainty.
In other words, a comprehensive prediction explanation should consist of \emph{label explanation} with important words and \emph{uncertainty explanation} with uncertain words.

The goal of this work is to demonstrate \textbf{the benefits of comprehensive explanations and the necessity of including uncertainty explanations}. 
In the empirical study, we adopt two explanation methods, Leave-one-out \citep{li2016understanding} and Sampling Shapley \citep{strumbelj2010efficient}, to explain two pre-trained language models, BERT \citep{devlin-etal-2019-bert} and RoBERTa \citep{liu2019roberta} on three tasks.
Experiments show the effectiveness of the two methods in identifying uncertain words for explaining model predictive uncertainty. 
Besides, human evaluations illustrate the indispensability of uncertainty explanations in helping humans understand model prediction behavior.

%% file: relate.tex
\section{Related Work}
\label{sec:relate}
The problem of predictive uncertainty estimation has been well studied \citep{kuleshov2015calibrated, gal2016dropout, pereyra2017regularizing, kumar2018trainable, liu2020simple, kong-etal-2020-calibrated, jiang2021can}. 
However, little is known about what causes predictive uncertainty. 
Extensive literatures on model explanations focus on explaining model predicted labels, while ignoring predictive uncertainty \citep{ribeiro2016should, lundberg2017unified, sundararajan2017axiomatic, chen-etal-2020-generating-hierarchical, chen-etal-2021-explaining}. 
However, explaining predictive uncertainty is an important complement to explaining predicted labels for improving model trustworthiness \citep{antoran2020getting, perez2022attribution}.

There is limited work on studying the source of predictive uncertainty in NLP. 
For example, previous works on explaining uncertainty estimates mainly focus on tabular and image data \citep{antoran2020getting, ley2021diverse, perez2022attribution}. 
\citet{feng-etal-2018-pathologies} observed that prediction confidence increases with input reduction, while focusing on model pathologies as reduced inputs lack predictive information. 
Differently, we focus on identifying uncertain words in inputs for explaining model predictive uncertainty. 
To the best of our knowledge, this is the first work on explaining predictive uncertainty of pre-trained language models in NLP.

%% file: method.tex
\section{Explaining Predictive Uncertainty}
\label{sec:method}
In this work, we consider models that are calibrated, such that their prediction confidence is aligned with their prediction probability.
Let $f(\cdot)$ denote a model. 
Given an input $\vec{x}=[{x}_{1}, \dots, {x}_{N}]$ consisting of $N$ words, the model prediction probabilities on $\vec{x}$ over classes are $[f_{1}(\vec{x}), \dots, f_{C}(\vec{x})]$, where $f_{c}(\vec{x})=P(y=c \mid \vec{x})$ and $C$ is the total number of classes. 
As model $f$ is calibrated, the probability on the predicted class $\hat{y}$, i.e. $f_{\hat{y}}(\vec{x})$, represents the model prediction confidence on this label. 
As prediction confidence and predictive uncertainty are negative correlated (higher confidence implies lower uncertainty), we explain model predictive uncertainty by answering the question: \textit{What drags model prediction confidence down?}
We answer the question based on a simple observation on model prediction explanations \citep{ribeiro2016should, li2016understanding, lundberg2017unified}.

When a feature is identified with negative contribution, removing it can improve model prediction confidence, as shown in \autoref{fig:illustrations}. 
Similar to the definition of prediction explanation, we consider this feature explains predictive uncertainty. 
Furthermore, given a ranking of input word contributions produced by an explanation method, we name top-ranked words as \emph{important words}, explaining model predicted labels; and bottom words (with negative contributions) as \emph{uncertain words}, explaining model predictive uncertainty. 
In other words, a comprehensive prediction explanation should consist of \emph{label explanation} with important words and \emph{uncertainty explanation} with uncertain words. 
As mentioned before, the goal of this study is to demonstrate the benefits of comprehensive explanations and the necessity of including uncertainty explanations.
In this work, we focus on extracting uncertain words from \emph{existing} explanation methods, with the expectation of stimulating further research on explaining predictive uncertainty in NLP.

\subsection{Explanation Methods}
With the previous discussion, we adopt two perturbation-based explanation methods, Leave-one-out \citep{li2016understanding} and Sampling Shapley \citep{strumbelj2010efficient}, for uncertainty explanations. 
Other explanation methods can be easily adapted to explaining predictive uncertainty. 
\paragraph{Leave-one-out (LOO).} 
This method evaluates the effect of each word on model prediction by leaving it out and observing the output probability change on the predicted class. 
We define a contribution score for each word as 
\begin{equation}
	\label{eq:loo}
	S_{i} = f_{\hat{y}}(\vec{x}) - f_{\hat{y}}(\vec{x}_{\setminus i}),
\end{equation}
where $\vec{x}_{\setminus i}$ denotes the input with the word $\vec{x}_{i}$ removed. 
The contribution score $S_{i}$ quantifies how much the model prediction confidence decreases when $\vec{x}_{i}$ is left out. 
\paragraph{Sampling Shapley (SS).} 
% Leave-one-out is simple but ignores coalitions between words when quantifying their contributions. 
% The Shapley value \citep{shapley1953value} stems from coalitional game theory provides an axiomatic solution to attribute the contribution of each word in a fair way. 
% However, the exponential complexity ($O(2^N)$) of computing Shapley value is intractable. 
% Sampling Shapley \citep{strumbelj2010efficient} provides a solvable approximation to Shapley value via sampling. 
This method computes feature contributions in a more sophisticated way by considering coalitions between words. 
Specifically, for a word $\vec{x}_i$, its contribution score is computed as 
\begin{equation}
		\label{eq:shapley}
		S_{i} = \frac{1}{M}\sum_{m=1}^{M} f_{\hat{y}}(\vec{x}_{\setminus i}^{(m)} \cup \{{x}_i\}) - f_{\hat{y}}(\vec{x}_{\setminus i}^{(m)}),
\end{equation}
where $M$ is the number of samples, and $\vec{x}_{\setminus i}^{(m)} \subseteq \vec{x}_{\setminus i}$ contains a subset of words in $\vec{x}_{\setminus i}$. 
The contribution score quantifies the overall contribution of the word $\vec{x}_{i}$ to the predicted label over $M$ ensembles. 
In experiments, we set $M=200$.

For each prediction, both methods produce an explanation with input word contributions, from which we extract important and uncertain words as label and uncertainty explanations respectively.

%% file: exp.tex
\section{Setup }
\label{sec:setup}
\paragraph{Models and datasets.}
We evaluate two pre-trained language models, BERT \citep{devlin-etal-2019-bert} and RoBERTa \citep{liu2019roberta}, on three tasks, including sentiment analysis, toxic comments detection and political bias classification. 
We utilize the IMDB \citep{maas2011learning} dataset for sentiment analysis, Wikipedia Toxicity Corpus (Toxics) \citep{wulczyn2017ex} for toxic comments detection, and Senator Tweets (Politics) \footnote{\url{https://huggingface.co/datasets/m-newhauser/senator-tweets}{}} for political bias classification. 
More details about the models and datasets are in Appendix \ref{sec:model_data}.

\paragraph{Posterior calibration.}
We follow \citet{desai-durrett-2020-calibration} and calibrate the models on each dataset via temperature scaling \citep{guo2017calibration}, so that their output probabilities on predicted labels well represent prediction confidence. 
More details of model calibration are in Appendix \ref{sec:post_cal}. 
% The calibration results are reported in \autoref{tab:calibrate}. 
% We apply temperature scaling to correct model outputs in the following experiments. 

\section{Experiments}
\label{sec:exp}
In our experiments, we focus on the three research questions: (1) How effectively existing model explanation methods can identify uncertain words? (2) What insights we can obtain from uncertainty explanations in addition to label explanations? (3) Whether users appreciate uncertainty explanations in understanding model prediction behavior?
\subsection{Quantitative Evaluation}
\label{sec:quan_eval}
For each dataset, we randomly select 1000 test examples and generate explanations for model predictions on them (see visualizations in \autoref{tab:exp}). 
The following two results answer the research question (1) and (2) respectively.

\begin{figure}[t]
	\centering
	\subfigure[BERT, IMDB]{
		\label{fig:bert_imdb}
		\includegraphics[width=0.23\textwidth]{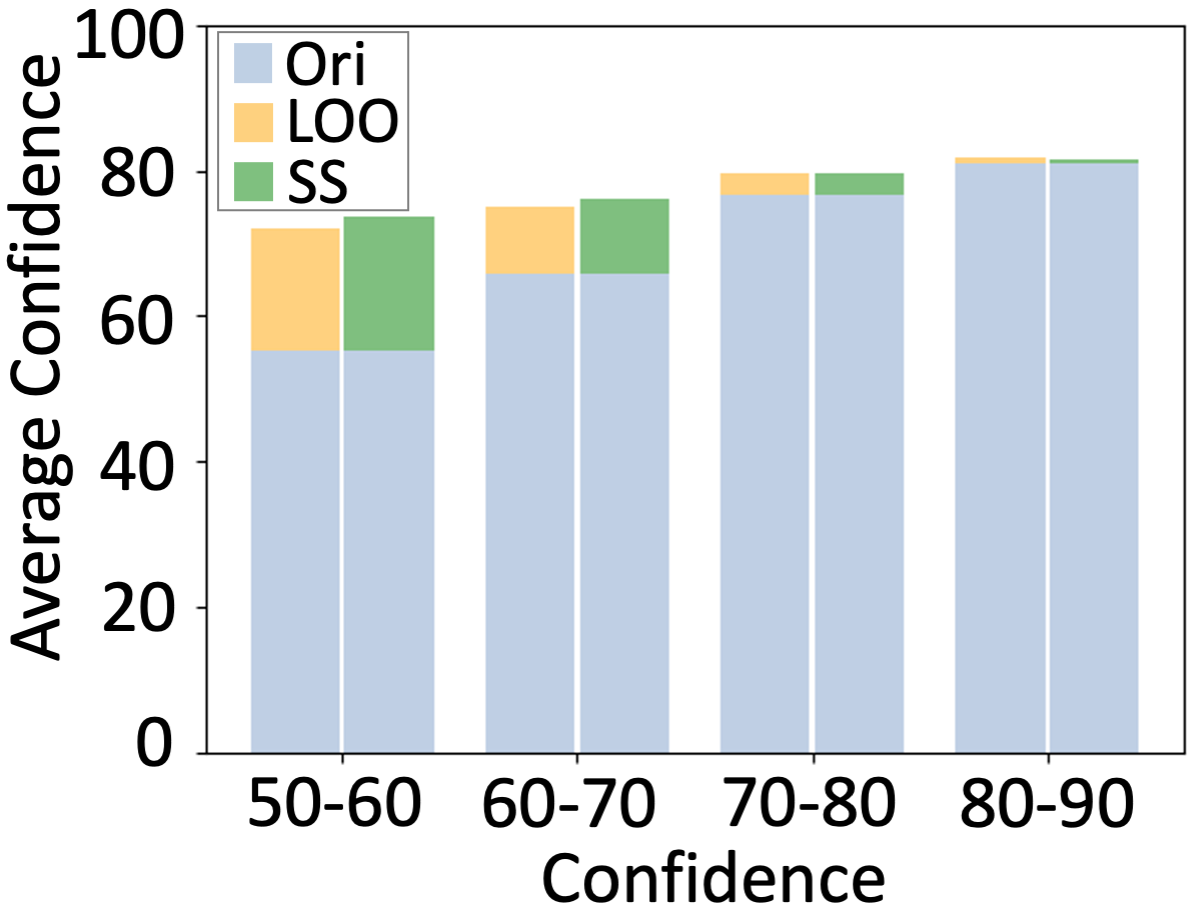}}
	\subfigure[RoBERTa, IMDB]{
		\label{fig:roberta_imdb}
		\includegraphics[width=0.23\textwidth]{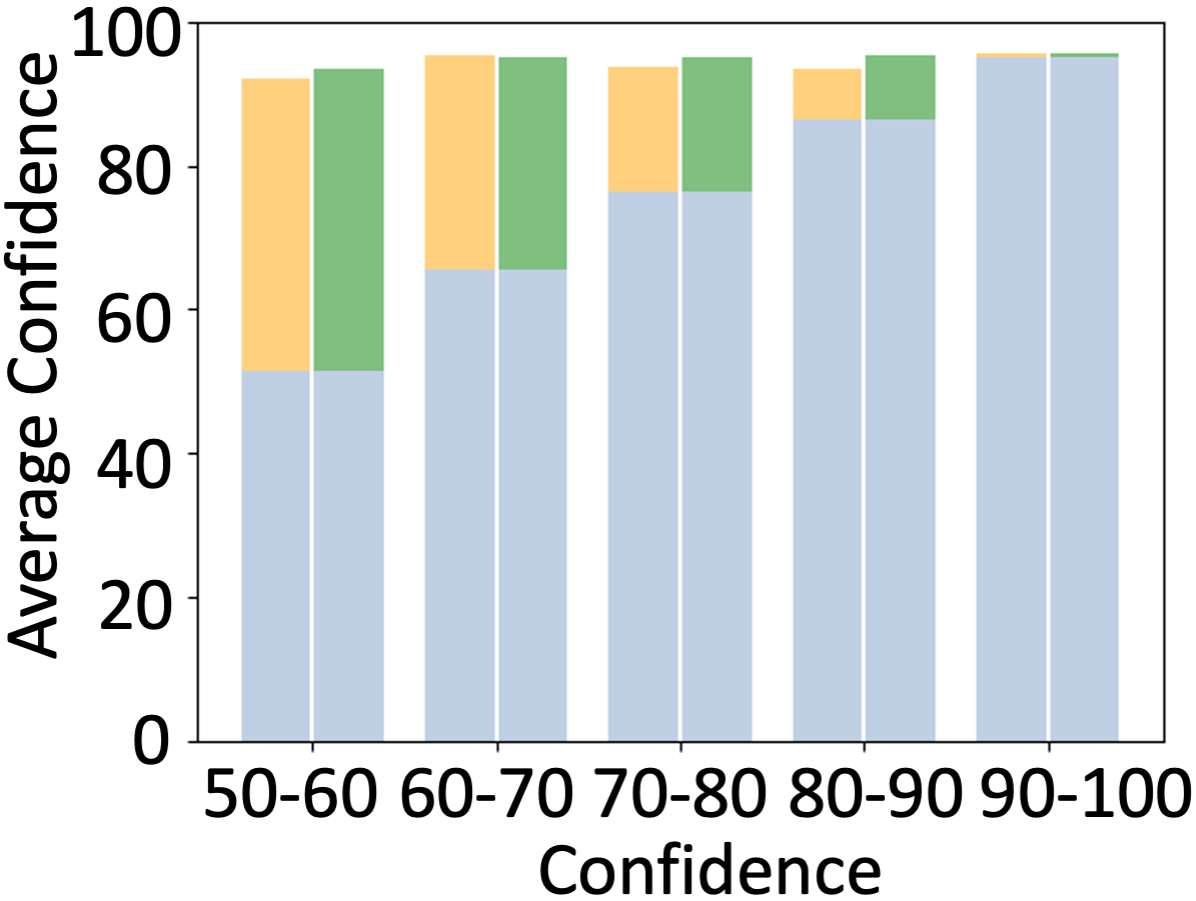}}
	\subfigure[BERT, Toxics]{
		\label{fig:bert_wiki}
		\includegraphics[width=0.23\textwidth]{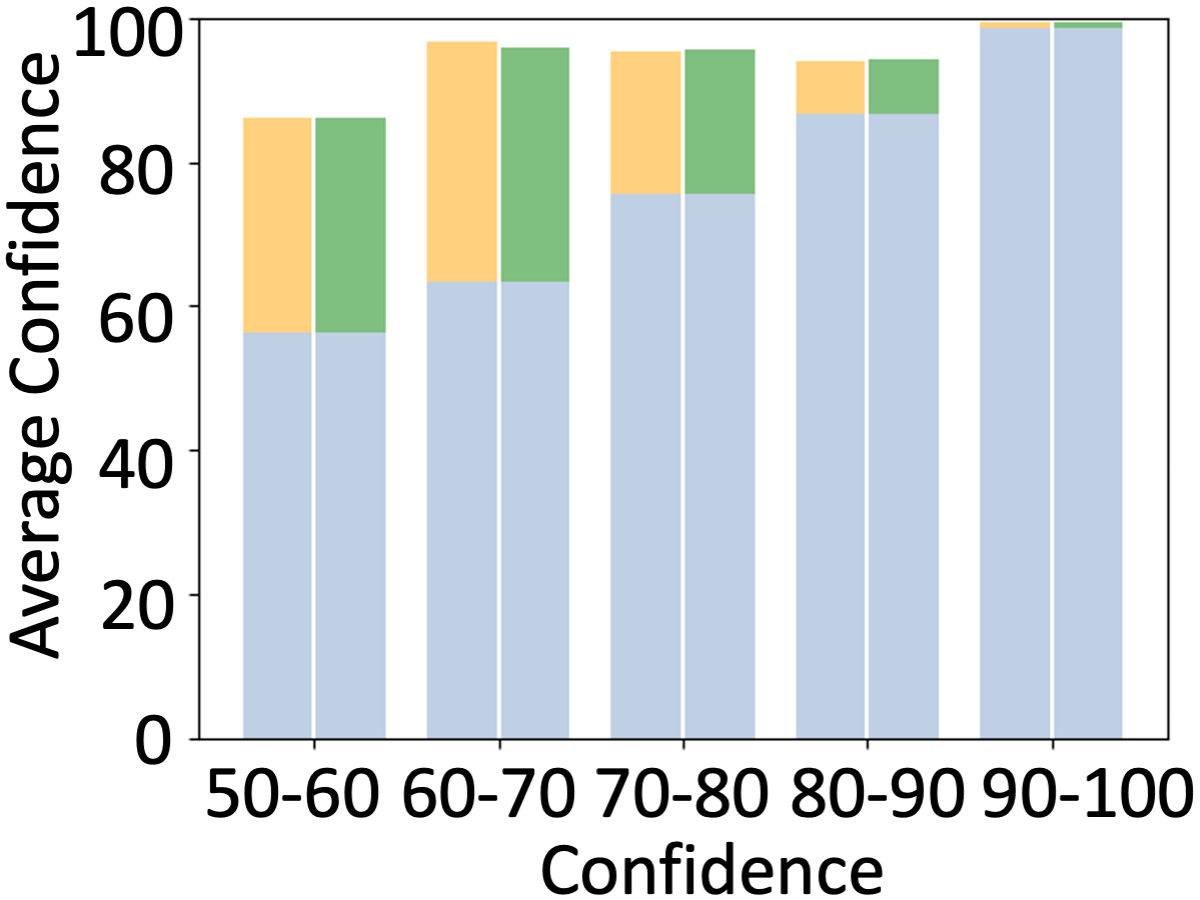}}
	\subfigure[RoBERTa, Toxics]{
		\label{fig:roberta_wiki}
		\includegraphics[width=0.23\textwidth]{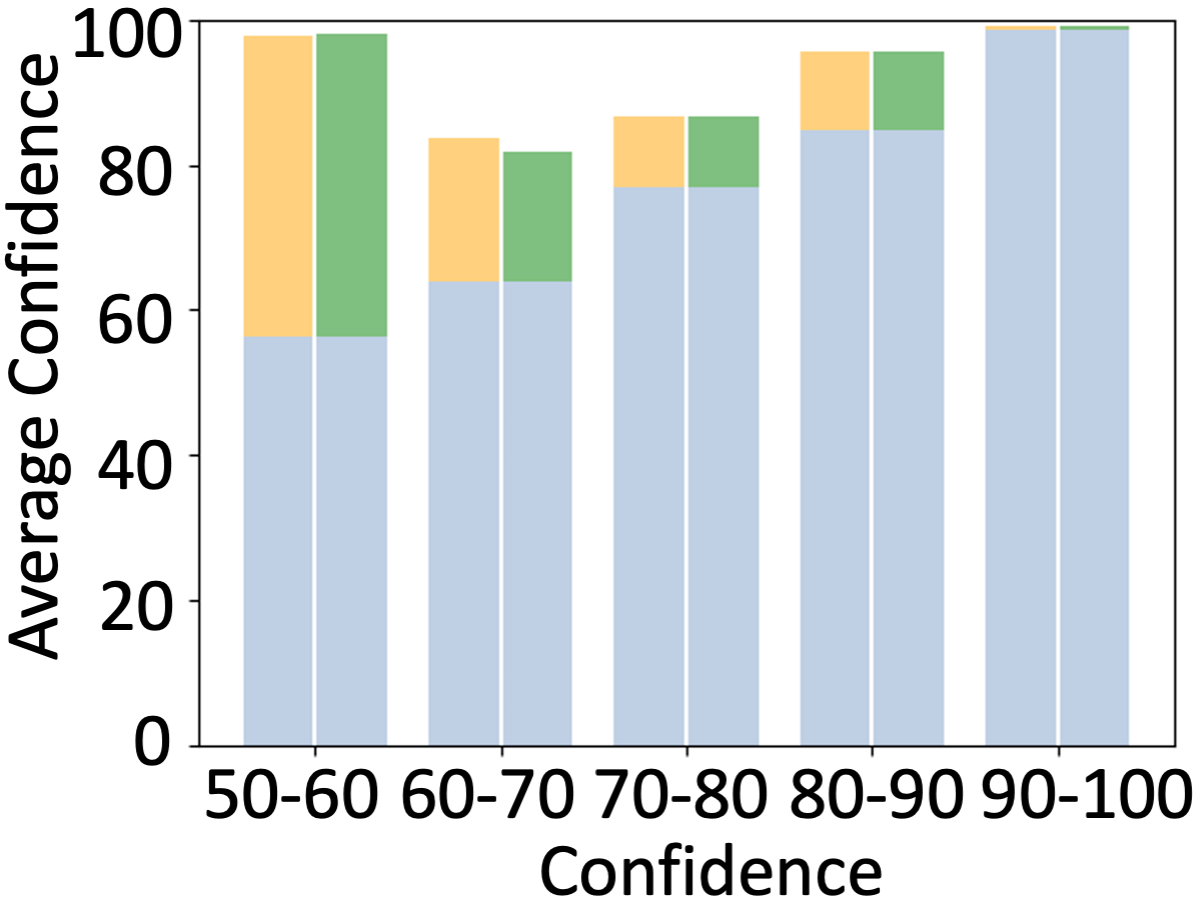}}
	\subfigure[BERT, Politics]{
		\label{fig:bert_senator}
		\includegraphics[width=0.23\textwidth]{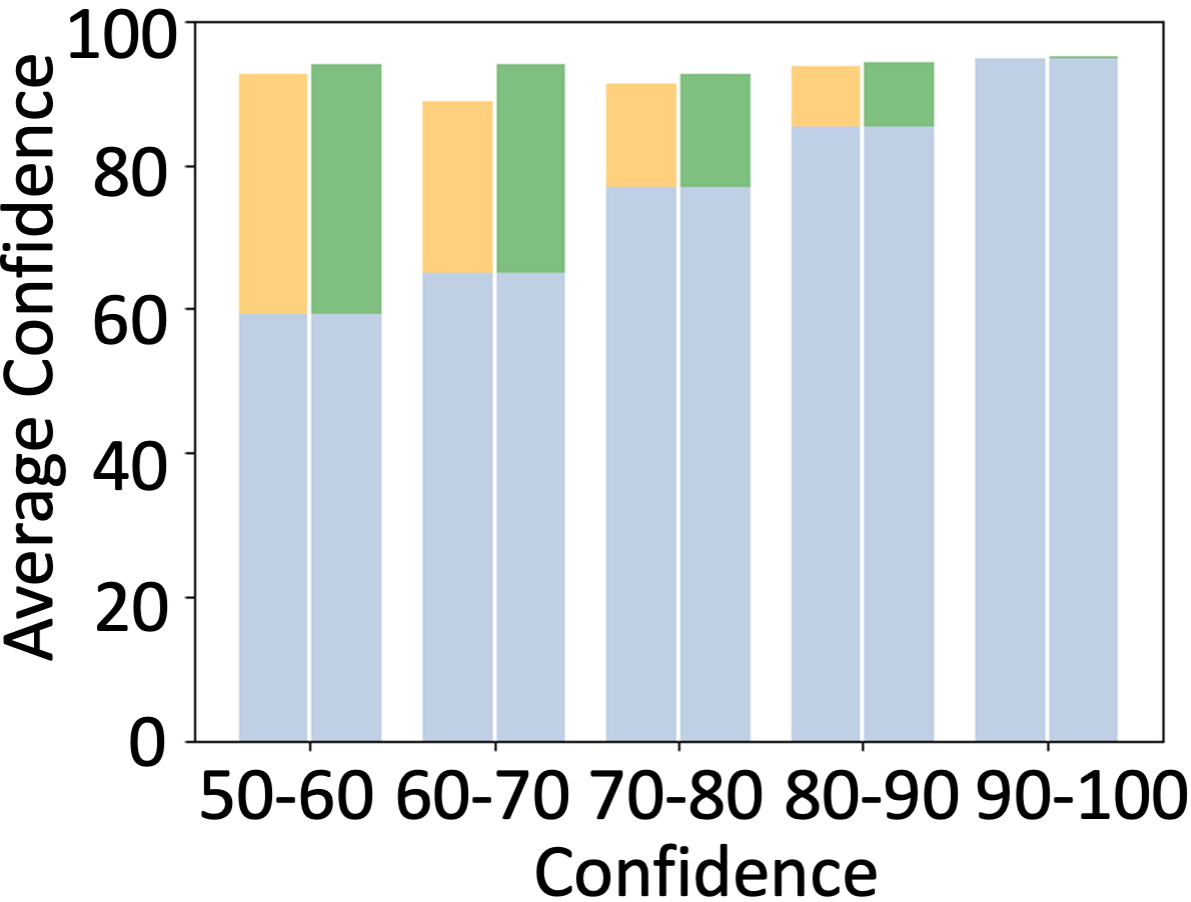}}
	\subfigure[RoBERTa, Politics]{
		\label{fig:roberta_senator}
		\includegraphics[width=0.23\textwidth]{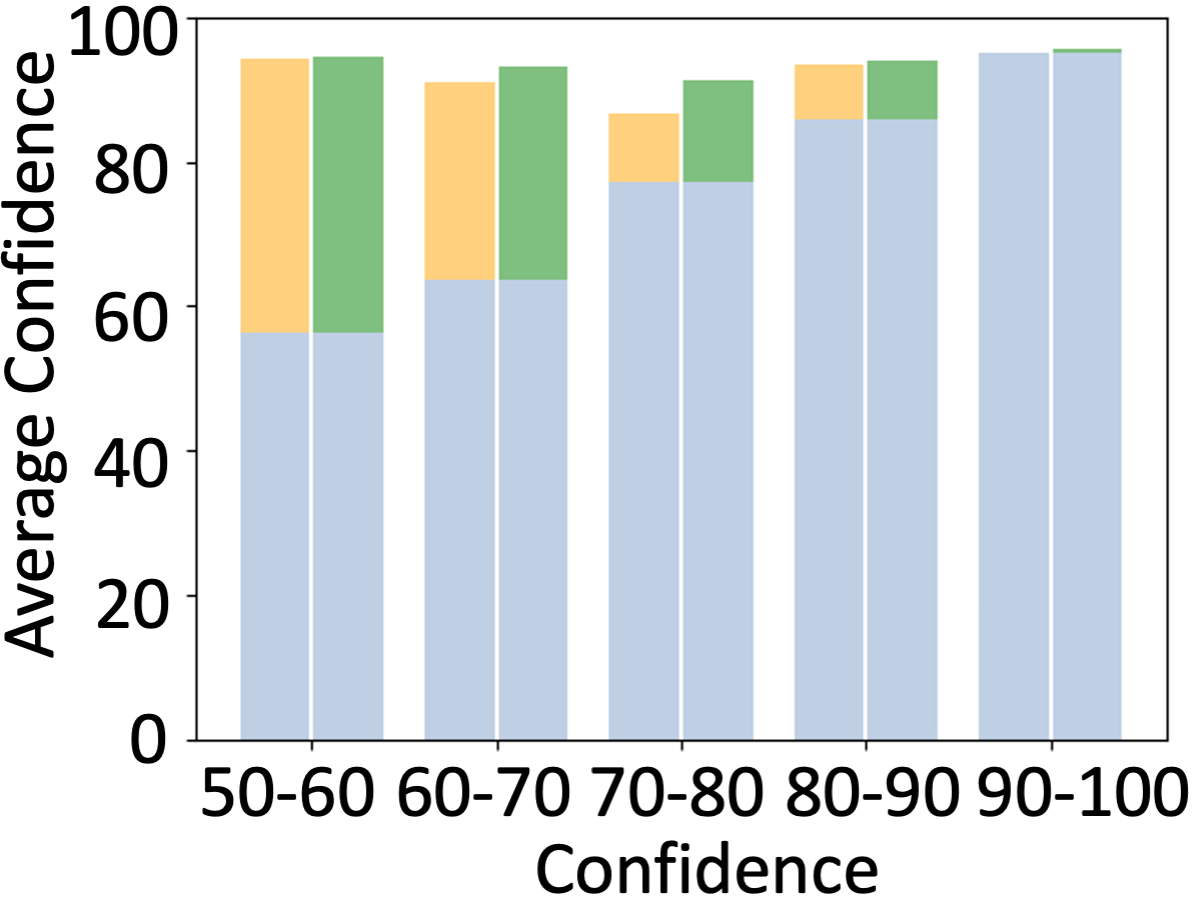}}
	\caption{Average confidence (\%) changes with uncertain words removed. X-axis shows different bins of original confidence. Ori: original confidence; LOO: Leave-one-out; SS: Sampling Shapley.}
	\label{fig:ave_conf}
\end{figure}

\paragraph{Existing model explanation methods effectively identify uncertain words that limit model prediction confidence.}
We extract top $k$ uncertain words identified by model explanations and remove them from inputs and then compute the average prediction confidence change in each bin of original confidence. 
We empirically set $k=10$ for IMDB and $k=5$ for Toxics and Politics based on their average sentence lengths in \autoref{tab:datasets}. 
\autoref{fig:ave_conf} shows that both LOO and SS capture uncertain words that limit prediction confidence. 
Overall, SS performs better than LOO in identifying uncertain words. 

\begin{figure}[t]
	\centering
	\includegraphics[width=0.47\textwidth]{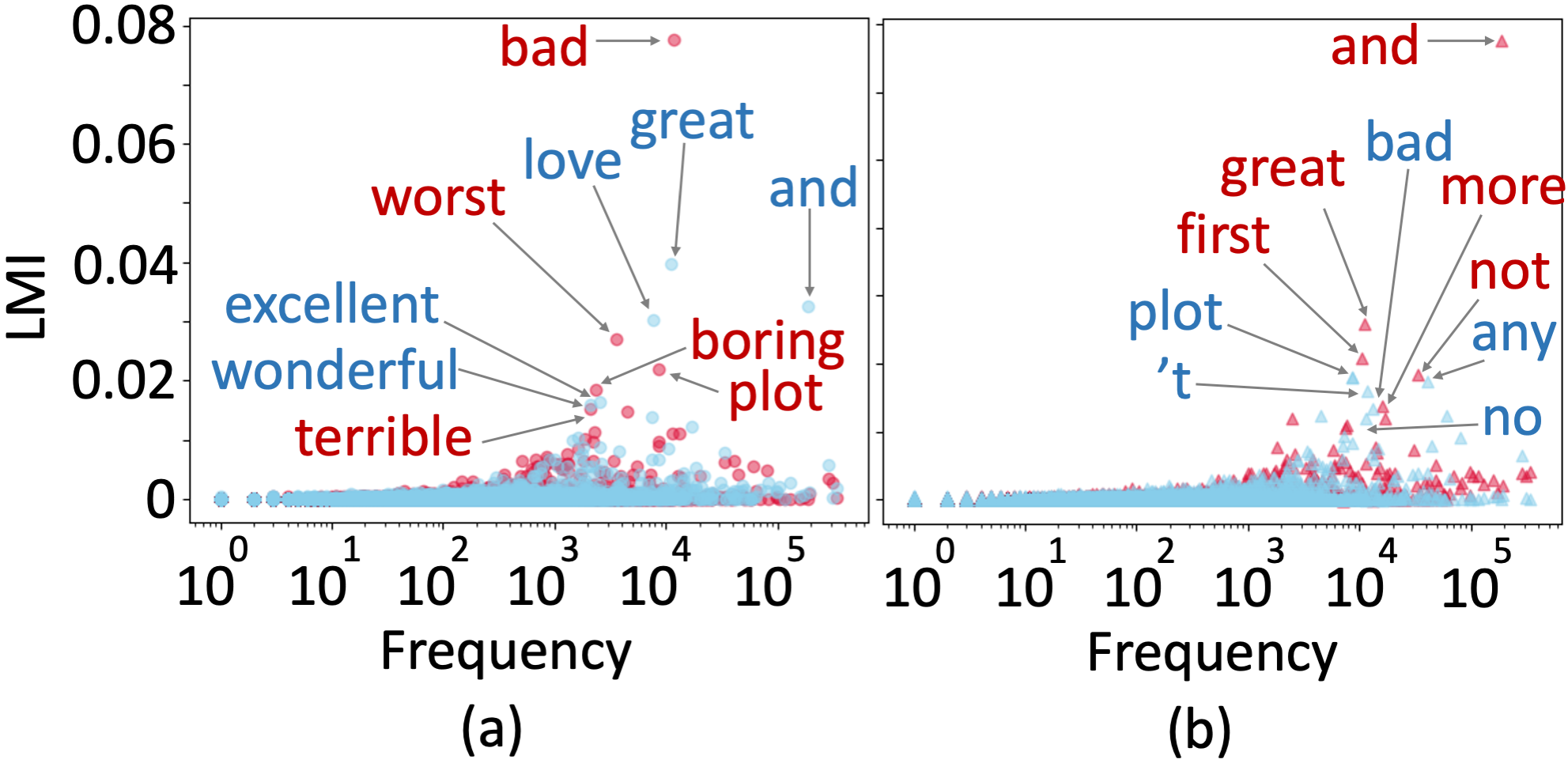}
	\caption{\label{fig:lmi_dist} LMI distributions based on important words (a) and uncertain words (b). The x-axis represents word frequency in the vocabulary built on the IMDB dataset. We use blue and red colors to distinguish features associated with the \textsc{positive} and \textsc{negative} labels respectively. Top 5 tokens in each distribution are pointed out.}
\end{figure}

\paragraph{Important words and negations can result in uncertain predictions.}
We analyze feature statistics of model explanations via local mutual information (LMI) \citep{schuster-etal-2019-towards, du-etal-2021-towards}. 
LMI quantifies the association between a feature (an important/uncertain word) and a prediction label in model explanations \citep{chen2022pathologies}. 
The details of computing LMI are in Appendix \ref{sec:lmi}. 
We analyze explanations generated by SS for RoBERTa on the IMDB dataset. 
Figure \ref{fig:lmi_dist} shows LMI distributions based on important and uncertain words in explanations respectively. 
Some important words for model predictions on a specific label (e.g., \texttt{great} for \textsc{positive}, \texttt{bad} for \textsc{negative} in (a)) become uncertain words for the other label in (b). 
This indicates models may get confused by important words corresponding to different labels in inputs. 
Besides, negation words (e.g., \texttt{not}, \texttt{no}) pointed out in (b) are not shown in (a), which means they may not be used by models for making predictions but can highly cause model predictive uncertainty. 
We observe similar results on other datasets in \autoref{tab:top_fea}. 

\input{tables/tab-human-eval}

\subsection{Human Evaluation}
\label{sec:human_eval}
To answer the research question (3), we conduct human evaluation on both important and uncertain words in model explanations through the Amazon Mechanical Turk (AMT). 
The details of human evaluation are in Appendix \ref{sec:amt_eval}. 
The following two observations illustrate the effectiveness and indispensability of uncertainty explanations.

\paragraph{Humans perform better on understanding uncertainty explanations than label explanations.}
First, we provide inputs with important words highlighted and ask evaluators to guess model prediction labels. 
Then we show model predictions with confidence and ask evaluators whether removing uncertain words can improve prediction confidence or not. 
\autoref{tab:human_eval} shows the results of human performance on predicting model prediction labels and confidence change. 
Overall, humans have better performance on understanding model predictive uncertainty based on uncertain words. 
This indicates the effectiveness of uncertainty explanations in helping users understand model predictions. 
Besides, SS produces more understandable explanations to humans than LOO. 
This is also reflected by the evaluation results where evaluators score (from 1-5) the quality of explanations with the average values 3.7 and 4.0 for LOO and SS respectively.

\paragraph{Humans prefer to see uncertainty explanations in addition to label explanations.}
We ask evaluators to vote whether they want to include uncertainty explanations in addition to label explanations for understanding model decision making. 
Most ($71\%$) evaluators prefer to see uncertainty explanations. 
Besides, evaluators mark $72.6\%$ of uncertainty explanations identify the words that largely limit model prediction confidence. 
This implies that uncertainty explanations are indispensable to explaining model prediction behavior.

%% file: tables/tab-human-eval.tex
\begin{table}[t]
	\small
	\centering
	\begin{tabular}{P{1.1cm}P{1.0cm}P{0.6cm}P{0.6cm}P{0.6cm}P{0.6cm}}
		\toprule
		\multirow{2}{*}{Model} & \multirow{2}{*}{Dataset} & \multicolumn{2}{c}{LOO} & \multicolumn{2}{c}{SS} \\
		\cmidrule(lr){3-4} \cmidrule(lr){5-6}
		 &  & Label & Unc & Label & Unc \\
		\midrule
		\multirow{3}{*}{BERT} & IMDB & 63.33 & \textbf{86.67} & 67.50 & \textbf{87.50} \\ 
		& Toxics & 60.00 & \textbf{86.67} & 86.67 & \textbf{90.00} \\
		& Politics & 66.67 & \textbf{83.33} & 82.50 & 75.00 \\
		\midrule
		\multirow{3}{*}{RoBERTa} & IMDB & 66.67 & \textbf{83.33} & 80.00 & \textbf{86.67} \\ 
	    & Toxics & 80.00 & \textbf{83.33} & 80.00 & \textbf{86.67} \\
		& Politics & 63.33 & 60.00 & 75.00 & 65.00 \\
		\bottomrule
	\end{tabular}
	\caption{Human prediction performance (\%) on label explanations (Label) and uncertainty explanations (Unc).}
	\label{tab:human_eval}
\end{table}

%% file: con.tex
\section{Conclusion}
\label{sec:con}
In this paper, we propose to explain model prediction uncertainty by extracting uncertain words from existing model explanations. 
We adopt two explanation methods to explain BERT and RoBERTa on three tasks. 
% , including sentiment analysis, toxic comments detection, and political bias classification. 
Experiments show the effectiveness of uncertainty explanations in explaining models and helping humans understand model predictions. 

%% file: appendix.tex
\section{Supplement of Experiments}
\label{sec:sup_exp}

\subsection{Models and Datasets}
\label{sec:model_data}
We adopt the pretrained BERT-base and RoBERTa-base models from Hugging Face\footnote{\url{https://github.com/huggingface/transformers}{}}. 
For sentiment analysis, we utilize the IMDB \citep{maas2011learning} dataset which contains positive and negative movie reviews. 
For toxic comments detection, we test on the Wikipedia Toxicity Corpus (Toxics) \citep{wulczyn2017ex}. 
The task is to detect whether a comment is toxic or nontoxic. 
For political bias classification, we adopt the Senator Tweets dataset (Politics) \footnote{\url{https://huggingface.co/datasets/m-newhauser/senator-tweets}{}}, which collects all tweets made by US senators during 2021-2022. 
The task is to recognize the political bias of each tweet as Democratic or Republican. 
All datasets are in English. 
Table \ref{tab:datasets} shows the statistics of the datasets. 
We fine-tune the models on the three datasets and report their prediction performance in Table \ref{tab:ori-acc}.  

\input{tables/tab-datasets}
\input{tables/tab-ori-acc}

We implement the models in PyTorch 3.7. The numbers of parameters in the BERT and RoBERTa models are 109484547 and 124647170 respectively. We manually set hyperparameters as: learning rate is $1e-5$, maximum sequence length is $256$, maximum gradient norm is $1$, and batch size is $8$. All experiments were performed on a single NVidia GTX 1080 GPU. The corresponding validation accuracy for each reported test accuracy is in \autoref{tab:val_acc}. The time for training each model on each dataset is in \autoref{tab:run_time}. All training and evaluation are based on one run.

\begin{table}[tbh]
	\centering
	\begin{tabular}{cccc}
		\toprule
		Models & IMDB & Toxics & Politics \\
		\midrule
		BERT  & 91.76 & 96.83 & 91.44 \\
		RoBERTa & 93.30 & 96.69 & 91.53 \\
		\bottomrule
	\end{tabular}
	\caption{Validation accuracy (\%) for each reported test accuracy.}
	\label{tab:val_acc}
\end{table}

\begin{table}[tbh]
	\centering
	\begin{tabular}{cccc}
		\toprule
		Models & IMDB & Toxics & Politics \\
		\midrule
		BERT  & 856.43 & 3254.33 & 1483.65 \\
		RoBERTa & 912.47 & 3467.39 & 1646.12 \\
		\bottomrule
	\end{tabular}
	\caption{The average runtime (s/epoch) of each model on each in-domain dataset.}
	\label{tab:run_time}
\end{table}

\subsection{Posterior Calibration}
\label{sec:post_cal}
A common way of measuring predictive uncertainty is by calibrating model outputs with the true correctness likelihood, so that the predictive probabilities well represent the confidence of model predictions being correct \citep{guo2017calibration, kong-etal-2020-calibrated, desai-durrett-2020-calibration, zhao2021calibrate}. 
Lower prediction confidence indicates higher uncertainty \citep{xu-etal-2020-understanding-neural, jiang2021can}. 
We follow the post-calibration methods and adopt the temperature scaling \citep{guo2017calibration, zhao2021calibrate} to calibrate the pre-trained language models (BERT and RoBERTa) in our experiments. 

Specifically, we use the development set to learn a temperature $T$ which corrects model output probabilities by dividing non-normalized logits before the softmax function. 
Then the learned $T$ is applied to modify model outputs on the test set. 
In experiments, we linearly search for an optimal temperature $T$ between $[0, 10]$ with a granularity of 0.01, which empirically performs well. 
We evaluate model calibration with Expected Calibration Error (ECE) \citep{guo2017calibration}. 
The ECE measures the difference between prediction confidence and accuracy, i.e. 
\begin{equation}
	\label{eq:ece}
	ECE = \sum_{k=1}^{K}\frac{\lvert B_k\rvert}{n} \lvert acc(B_k) - conf(B_k) \rvert,
\end{equation}
where the total $n$ predictions are partitioned into $K$ equally-spaced bins, $B_k$ represents the predictions fall into the $k$th bin, $acc(\cdot)$ and $conf(\cdot)$ compute the average accuracy and confidence in each bin respectively. 
For a perfect calibration, $acc(B_k)=conf(B_k)$, $k \in \{1, \dots, K\}$. 
In this work, we set $K=10$. 
We report the learned temperature scalars and ECEs before and after calibration in \autoref{tab:calibrate}. 
Temperature scaling performs effectively in decreasing model calibration errors. 
This enables us to further explain prediction uncertainty based on calibrated confidence. 
We apply temperature scaling to correct model outputs in experiments.

\input{tables/tab-calibrate}

\input{tables/tab-top}

\subsection{Local Mutual Information}
\label{sec:lmi}
To understand which features contribute to model predictions and which features cause prediction uncertainty, we follow \citep{schuster-etal-2019-towards, du-etal-2021-towards, chen2022pathologies} and analyze feature statistics of model explanations via local mutual information (LMI). 
LMI quantifies the association between a feature and a prediction label in model explanations. 
We compute LMI based on top $5$ important and uncertain words in prediction and uncertainty explanations respectively. 
Specifically, for each group of features, we can get a set of unique features, $E=\{e\}$. 
The LMI between a feature $e$ and a prediction label $y$ is 
\begin{equation}
	\label{eq:lmi}
	\text{LMI}(e, y)=p(e, y)\cdot \log \left(\frac{p(y \mid e)}{p(y)}\right),
\end{equation}
where $p(y \mid e)=\frac{count(e, y)}{count(e)}$, $p(y)=\frac{count(y)}{\lvert E \rvert}$, $p(e, y)=\frac{count(e, y)}{\lvert E \rvert}$, and $\lvert E \rvert$ is the number of occurrences of all features in $E$. 
Then we can get a distribution of LMI over all tokens in the vocabulary ($\{w\}$) built on the dataset, i.e.
\begin{equation}
	\label{eq:p_lmi}
	P_{\text{LMI}}(w, y) = 
	\begin{cases}
		\text{LMI}(w, y) & \text{if token $w \in E$}\\
		0 & \text{else}
	\end{cases}
\end{equation}
We normalize the LMI distribution by dividing each value with the sum of all values. 
Table \ref{tab:top_fea} records top 10 tokens in different LMI distributions of model explanations. 
% Leave-one-out captures more noisy tokens (e.g., special tokens, punctuations) than Sampling Shapley. 
% Both BERT and RoBERTa focus on some task-irrelevant features (e.g., stop words) to make predictions, which reveals the problem of model prediction behavior. 
% We leave this problem to our future work. 

\subsection{Human Evaluation}
\label{sec:amt_eval}
We conduct human evaluation on both important and uncertain words in model explanations through the Amazon Mechanical Turk (AMT). 
For each dataset, we randomly select 30 test examples to generate explanations for each pre-trained language model. 
Each explanation (with 2-3 important and uncertain words extracted respectively) is assessed by 5 workers. 
We pay the workers $\$0.3$ for assessing each explanation. 
We have collected 900 annotations in total. 

\begin{figure*}[t]
  \centering
  \includegraphics[width=0.8\textwidth]{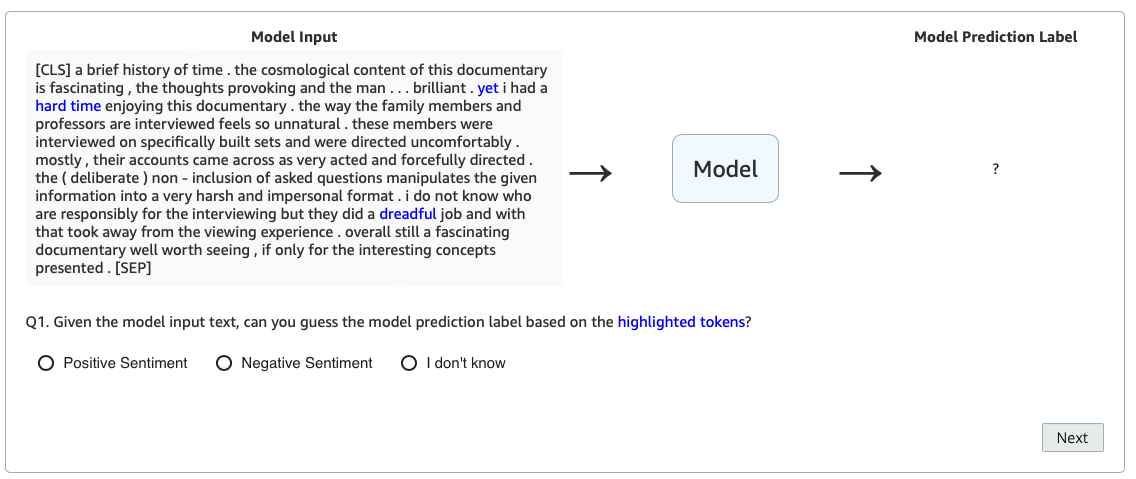}
  \caption{\label{fig:human_eval_imp} Interface of human evaluation on important words highlighted in blue color.}
\end{figure*}

\begin{figure*}[t]
  \centering
  \includegraphics[width=0.8\textwidth]{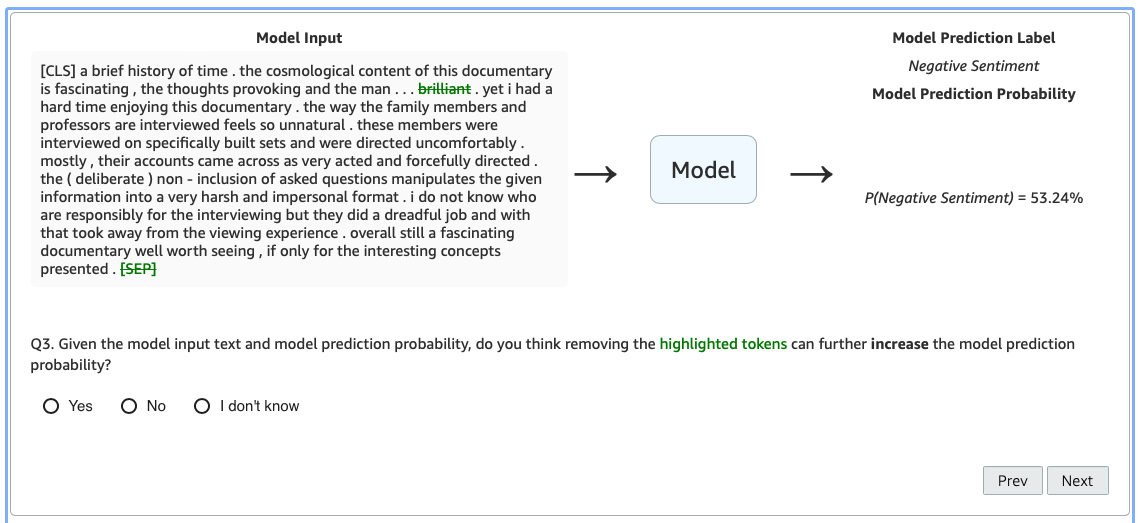}
  \caption{\label{fig:human_eval_unc} Interface of human evaluation on uncertain words highlighted in green color.}
\end{figure*}

For each explanation, we ask the worker to answer the following 5 questions:
\begin{enumerate}
    \item \textbf{Prediction on label explanations (multiple choices)}: Given the model input text, can you guess the model prediction label based on the highlighted tokens? 
    \item \textbf{Rating on label explanations (1-5 Liker scale)}: Given the model input text and model prediction label, how much do you think the highlighted tokens make sense to you?
    \item \textbf{Prediction on uncertainty explanations (multiple choices)}: Given the model input text and model prediction probability, do you think removing the highlighted tokens can further increase the model prediction probability or not?
    \item \textbf{Rating on uncertainty explanations (1-3 Liker scale)}: How much do you think the current model prediction probability could be changed by removing the highlighted tokens?
    \item \textbf{Comparison on label explanations and uncertainty explanations (multiple choices)}: Which type of model explanations can help you better understand the model prediction?
\end{enumerate}
\autoref{fig:human_eval_imp} and \autoref{fig:human_eval_unc} show the interfaces of human evaluation on Q1 and Q3 respectively. 

\subsection{Visualizations}
\label{sec:visual}
\autoref{tab:exp} shows visualizations of different model explanations with both important and uncertain words highlighted.

\input{tables/tab-exp}

%% file: tables/tab-datasets.tex
%  \begin{table*}[t] %tbph
%  	\centering
% %  	\small
%  	\begin{tabular}{cccccc}
%  		\toprule
%  		Datasets & \textit{L} & \textit{\#train} & \textit{\#dev} & \textit{\#test} & Label distribution \\
%  		\midrule
%  		IMDB & 231 & 20K & 5K & 25K & Positive: \textit{train}(10036), \textit{dev}(2414), \textit{test}(12535) \\
%  		 &  &  &  &  & Negative: \textit{train}(9956), \textit{dev}(2583), \textit{test}(12451) \\
%  		 \midrule
%  		Wiki & 68 & 96K & 32K & 32K & Toxic: \textit{train}(9245), \textit{dev}(3069), \textit{test}(3048) \\
%  		 &  &  &  &  & Nontoxic: \textit{train}(86447), \textit{dev}(29059), \textit{test}(28818) \\
%  		 \midrule
%  		Senat & 34 & 70K & 7.8K & 19K &  Democratic: \textit{train}(36222), \textit{dev}(3982), \textit{test}(10240) \\
%  		 &  &  &  &  & Republican: \textit{train}(33796), \textit{dev}(3789), \textit{test}(9189) \\
%  		\bottomrule
%  	\end{tabular}
%  	\caption{Summary statistics of the datasets, where \textit{L} is average sentence length, and \textit{\#} counts the number of examples in the \textit{train/dev/test} sets. For label distribution, the number of examples with a specific label in \textit{train/dev/test} is noted in bracket.}
%  	\label{tab:datasets}
%  \end{table*}

\begin{table*}[t] %tbph
 	\centering
 	\begin{tabular}{cccccc}
 		\toprule
 		Datasets & \textit{L} & \textit{\#train} & \textit{\#dev} & \textit{\#test} & Label distribution \\
 		\midrule
 		IMDB & 231 & 20K & 5K & 25K & Positive: \textit{train}(10036), \textit{dev}(2414), \textit{test}(12535) \\
 		 &  &  &  &  & Negative: \textit{train}(9956), \textit{dev}(2583), \textit{test}(12451) \\
 		 \midrule
 		Toxics & 68 & 96K & 32K & 32K & Toxic: \textit{train}(9245), \textit{dev}(3069), \textit{test}(3048) \\
 		 &  &  &  &  & Nontoxic: \textit{train}(86447), \textit{dev}(29059), \textit{test}(28818) \\
 		 \midrule
 		Politics & 34 & 70K & 7.8K & 19K &  Democratic: \textit{train}(36222), \textit{dev}(3982), \textit{test}(10240) \\
 		 &  &  &  &  & Republican: \textit{train}(33796), \textit{dev}(3789), \textit{test}(9189) \\
 		\bottomrule
 	\end{tabular}
 	\caption{Summary statistics of the datasets, where \textit{L} is average sentence length, and \textit{\#} counts the number of examples in the \textit{train/dev/test} sets. For label distribution, the number of examples with a specific label in \textit{train/dev/test} is noted in bracket.}
 	\label{tab:datasets}
\end{table*}

%% file: tables/tab-ori-acc.tex
\begin{table}[t] %tbph
% 	\small
	%   \setlength{\belowcaptionskip}{-5pt}
	\centering
	\begin{tabular}{cccc}
		\toprule
		Models & IMDB & Toxics & Politics \\
		\midrule
		BERT  & 91.29 & 96.96 & 91.20 \\
		RoBERTa & 93.36 & 96.75 & 91.32 \\
		\bottomrule
	\end{tabular}
	\caption{Prediction accuracy (\%) of different models on the test sets.}
	\label{tab:ori-acc}
\end{table}

%% file: tables/tab-calibrate.tex
\begin{table}[t] %tbph
	\centering
	\begin{tabular}{cccc}
		\toprule
		Models & IMDB & Toxics & Politics \\
		\midrule
		\textbf{BERT}: \\
		$T$ & 4.59 & 1.95 & 4.2 \\
		pre-ECE & 8.45 & 2.36 & 8.56 \\
		post-ECE & 2.85 & 0.89 & 3.83 \\
		\midrule
		\textbf{RoBERTa}: \\
		$T$ & 2.76 & 2.16 & 3.98 \\
		pre-ECE & 6.36 & 2.90 & 8.45 \\
		post-ECE & 2.50 & 1.13 & 4.29 \\
		\bottomrule
	\end{tabular}
	\caption{Posterior calibration results. $T$ is the learned temperature. pre-ECE and post-ECE represent the ECEs on test sets before and after calibration respectively.}
	\label{tab:calibrate}
\end{table}

%% file: tables/tab-top.tex
\begin{table*}[t]
	\small
	\centering
	\begin{tabular}{P{1cm}P{0.8cm}P{0.6cm}P{2.6cm}P{2.6cm}P{2.6cm}P{2.6cm}}
		\toprule
		\multirow{2}{*}{Model} & \multirow{2}{*}{Dataset} & \multirow{2}{*}{Label} & \multicolumn{2}{c}{Leave-one-out} & \multicolumn{2}{c}{Sampling Shapley} \\
		\cmidrule(lr){4-5} \cmidrule(lr){6-7}
		 &  &  & Important & Uncertain & Important & Uncertain \\
		\midrule
		\multirow{6}{*}{BERT} & \multirow{2}{*}{IMDB} & Pos & this great best film a good excellent and it wonderful & i movie this was to just the would but not & great best and excellent love wonderful good this very enjoyed & movie just would bad but nothing not could off plot \\   
		\rule{0pt}{3ex} 
	    & & Neg & this worst movie bad not but no terrible just nothing & but is and not the it a this 't great & bad worst this movie just boring terrible awful not nothing  & and great very it is good in not his seen \\
	    \rule{0pt}{5ex} 
	    & \multirow{2}{*}{Toxics} & Tox & you fuck hell fucking bullshit idiot dick suck stupid gay & the are so fuck of good have wow love for & you fuck hell gay fucking bullshit idiot dick suck stupid & the can in please if so certainly because know help \\
	    \rule{0pt}{3ex} 
	    & & NTox & please to i if not the of wikipedia can is thank & you your i the and a to please me is & please the can to if for in of thank use & you a i your me the my it van and \\
	    \rule{0pt}{5ex} 
	    & \multirow{2}{*}{Politics} & Dec & and to climate must this child our so the in & the and to of we in american is americans i & this must climate that health now today more every to  & the a is for american back ensure work and not \\
	    \rule{0pt}{3ex} 
	    & & Rep & democrats the is border bid great and communist inflation fox & to and the i this our my with in for & the a is bid and border for communist his fox & that this to more must you today every now your \\
		\midrule
		\multirow{6}{*}{RoBERTa} & \multirow{2}{*}{IMDB} & Pos & this best and great not but good I film is & the not I is a for this and no was & great and love excellent wonderful best very amazing brilliant perfect & any plot 't bad no nothing movie much never this \\   
		\rule{0pt}{3ex} 
	    & & Neg & this not bad worst boring just the even and no & bad 't not and plot butwas a to me & bad worst plot boring terrible nothing stupid much no waste & and great first not more special very love life moments \\
	    \rule{0pt}{5ex} 
	    & \multirow{2}{*}{Toxics} & Tox & you fuck stupid HELL suck Fuck YOU You fucking shit  & to but if or Go an reported thanks ipedia should & you fuck You stupid HELL suck Fuck YOU fucking shit & to for reported OF but the about in help need \\
	    \rule{0pt}{3ex} 
	    & & NTox & to the Please article of please for Thank and in & you your is I a Your vandal not my me & to the for article use in please Please of If & you your a me is my You vandal I are \\
	    \rule{0pt}{5ex} 
	    & \multirow{2}{*}{Politics} & Dec & and the to this in our a must for will & the americ an in to for is act this a & this and climate care child today that workers how families & americ is the will of back family they not would \\
	    \rule{0pt}{3ex} 
	    & & Rep & bid en is democr the border americ us great to & to the and our my i of in this a & americ border democr bid is spending great not inflation would & and this that to our it more my you families \\
		\bottomrule
	\end{tabular}
	\caption{Top 10 tokens in different LMI distributions of model explanations. Important: statistics of top salient words in explanations; Uncertain: statistics of bottom salient words in explanations; Pos: postive; Neg: negative; Tox: toxic; NTox: nontoxic; Dec: democratic; Rep: republican. \textcolor{red}{Warning: this table contains toxic tokens.}}
	\label{tab:top_fea}
\end{table*}

%% file: tables/tab-exp.tex
\begin{table*}[tbph]
	\small
	\centering
	\begin{tabular}{P{2.5cm}P{1.5cm}P{3.5cm}P{6.5cm}}
		\toprule
		Model/Dataset & Method & Prediction & Explanation \\
		\midrule
		BERT/IMDB & LOO & Negative (0.69 $\rightarrow$ 0.80) & i \hlc[cyan!25]{found} it to be a complete \hlc[cyan!50]{disappointment} . if i had of known this movie was going to be as stupid as it was , i would have stayed home and done something more entertaining ... the plot was a \hlc[red!50]{great} \hlc[red!25]{idea} , just could have been done in a much better way . \\
		\rule{0pt}{3ex}    
		RoBERTa/IMDB & SS & Positive (0.51 $\rightarrow$ 0.86) & \hlc[red!50]{Not} the \hlc[red!50]{best} of the Lone Star series, but it moves along \hlc[cyan!50]{quickly} \hlc[cyan!25]{with} good performances. Introduced as "Singin' Sandy" in the main title, John Wayne as a 'singing cowboy' isn't successful...  \\
		\rule{0pt}{3ex}    
		BERT/Toxics & LOO & Nontoxic (0.83 $\rightarrow$ 0.97) & oh , and i have a \hlc[cyan!25]{question} . why was the article on brad christian , a famous magician , deleted because of vandalism instead of simply restored ? i believe that many users on this site are biased towards magicians . i have come to the conclusion that wiki is a useless site that does nothing to help anyone . you are welcome to ban me longer , and i understand completely \hlc[cyan!50]{if} you do , but this site is the worst piece of \hlc[red!50]{garbage} i have ever found \hlc[red!25]{!} \\
		\rule{0pt}{3ex}    
		RoBERTa/Toxics & SS & Nontoxic (0.71 $\rightarrow$ 0.94) & You are so \hlc[red!25]{full} of \hlc[red!50]{shit}. First of all, you aren't an admin, and for the sake of this \hlc[cyan!25]{site} I hope you never will be. I know I will personally work against you if you ever decide to try for one. \hlc[cyan!50]{But} I digress as you are not an administrator, and especially since you have no access to checkuser, you cannot determine who is or is not a sockpuppet nor do you have the authorization to place a tag on a user page . \\
		\rule{0pt}{3ex}
		BERT/Politics & SS & Democratic (0.64 $\rightarrow$ 0.95) & fantastic news . star plastics was founded in ravenswood and is continuing to \hlc[cyan!50]{invest} in \hlc[red!50]{west} \hlc[red!25]{virginia} . this expansion will lead to economic development and growth in jackson county , and shows that wv is the \hlc[cyan!25]{perfect} place for companies large and small . \\
		\rule{0pt}{3ex}    
		RoBERTa/Politics & LOO & Republican (0.85 $\rightarrow$ 0.94) & \hlc[red!25]{happy} national day , taiwan. your commitment to \hlc[red!50]{democracy} and \hlc[cyan!25]{market} economics is an \hlc[cyan!50]{effective} model that can be relied upon to solve our collective problem . \\
		\bottomrule
	\end{tabular}
	\caption{Visualizations of prediction explanations for different models on different datasets, where top two important and uncertain words are highlighted in blue and red colors respectively. The prediction confidence changes are shown in brackets when the highlighted uncertain words are removed. LOO: Leave-one-out; SS: Sampling Shapley. \textcolor{red}{Warning: some examples may be offensive or upsetting.}}
	\label{tab:exp}
\end{table*}

%% file: main.bbl
\begin{thebibliography}{32}
\expandafter\ifx\csname natexlab\endcsname\relax\def\natexlab#1{#1}\fi

\bibitem[{Antor{\'a}n et~al.(2020)Antor{\'a}n, Bhatt, Adel, Weller, and
  Hern{\'a}ndez-Lobato}]{antoran2020getting}
Javier Antor{\'a}n, Umang Bhatt, Tameem Adel, Adrian Weller, and
  Jos{\'e}~Miguel Hern{\'a}ndez-Lobato. 2020.
\newblock Getting a clue: A method for explaining uncertainty estimates.
\newblock \emph{arXiv preprint arXiv:2006.06848}.

\bibitem[{Brown et~al.(2020)Brown, Mann, Ryder, Subbiah, Kaplan, Dhariwal,
  Neelakantan, Shyam, Sastry, Askell et~al.}]{brown2020language}
Tom~B Brown, Benjamin Mann, Nick Ryder, Melanie Subbiah, Jared Kaplan, Prafulla
  Dhariwal, Arvind Neelakantan, Pranav Shyam, Girish Sastry, Amanda Askell,
  et~al. 2020.
\newblock Language models are few-shot learners.
\newblock \emph{arXiv preprint arXiv:2005.14165}.

\bibitem[{Chen et~al.(2021)Chen, Feng, Ganhotra, Wan, Gunasekara, Joshi, and
  Ji}]{chen-etal-2021-explaining}
Hanjie Chen, Song Feng, Jatin Ganhotra, Hui Wan, Chulaka Gunasekara, Sachindra
  Joshi, and Yangfeng Ji. 2021.
\newblock \href {https://doi.org/10.18653/v1/2021.naacl-main.306} {Explaining
  neural network predictions on sentence pairs via learning word-group masks}.
\newblock In \emph{Proceedings of the 2021 Conference of the North American
  Chapter of the Association for Computational Linguistics: Human Language
  Technologies}, pages 3917--3930, Online. Association for Computational
  Linguistics.

\bibitem[{Chen et~al.(2020)Chen, Zheng, and
  Ji}]{chen-etal-2020-generating-hierarchical}
Hanjie Chen, Guangtao Zheng, and Yangfeng Ji. 2020.
\newblock \href {https://doi.org/10.18653/v1/2020.acl-main.494} {Generating
  hierarchical explanations on text classification via feature interaction
  detection}.
\newblock In \emph{Proceedings of the 58th Annual Meeting of the Association
  for Computational Linguistics}, pages 5578--5593, Online. Association for
  Computational Linguistics.

\bibitem[{Chen et~al.(2022)Chen, Zheng, Awadallah, and
  Ji}]{chen2022pathologies}
Hanjie Chen, Guoqing Zheng, Ahmed~Hassan Awadallah, and Yangfeng Ji. 2022.
\newblock Pathologies of pre-trained language models in few-shot fine-tuning.
\newblock \emph{arXiv preprint arXiv:2204.08039}.

\bibitem[{Desai and Durrett(2020)}]{desai-durrett-2020-calibration}
Shrey Desai and Greg Durrett. 2020.
\newblock \href {https://doi.org/10.18653/v1/2020.emnlp-main.21} {Calibration
  of pre-trained transformers}.
\newblock In \emph{Proceedings of the 2020 Conference on Empirical Methods in
  Natural Language Processing (EMNLP)}, pages 295--302, Online. Association for
  Computational Linguistics.

\bibitem[{Devlin et~al.(2019)Devlin, Chang, Lee, and
  Toutanova}]{devlin-etal-2019-bert}
Jacob Devlin, Ming-Wei Chang, Kenton Lee, and Kristina Toutanova. 2019.
\newblock \href {https://doi.org/10.18653/v1/N19-1423} {{BERT}: Pre-training of
  deep bidirectional transformers for language understanding}.
\newblock In \emph{Proceedings of the 2019 Conference of the North {A}merican
  Chapter of the Association for Computational Linguistics: Human Language
  Technologies, Volume 1 (Long and Short Papers)}, pages 4171--4186,
  Minneapolis, Minnesota. Association for Computational Linguistics.

\bibitem[{Du et~al.(2021)Du, Manjunatha, Jain, Deshpande, Dernoncourt, Gu, Sun,
  and Hu}]{du-etal-2021-towards}
Mengnan Du, Varun Manjunatha, Rajiv Jain, Ruchi Deshpande, Franck Dernoncourt,
  Jiuxiang Gu, Tong Sun, and Xia Hu. 2021.
\newblock \href {https://doi.org/10.18653/v1/2021.naacl-main.71} {Towards
  interpreting and mitigating shortcut learning behavior of {NLU} models}.
\newblock In \emph{Proceedings of the 2021 Conference of the North American
  Chapter of the Association for Computational Linguistics: Human Language
  Technologies}, pages 915--929, Online. Association for Computational
  Linguistics.

\bibitem[{Feng et~al.(2018)Feng, Wallace, Grissom~II, Iyyer, Rodriguez, and
  Boyd-Graber}]{feng-etal-2018-pathologies}
Shi Feng, Eric Wallace, Alvin Grissom~II, Mohit Iyyer, Pedro Rodriguez, and
  Jordan Boyd-Graber. 2018.
\newblock \href {https://doi.org/10.18653/v1/D18-1407} {Pathologies of neural
  models make interpretations difficult}.
\newblock In \emph{Proceedings of the 2018 Conference on Empirical Methods in
  Natural Language Processing}, pages 3719--3728, Brussels, Belgium.
  Association for Computational Linguistics.

\bibitem[{Gal and Ghahramani(2016)}]{gal2016dropout}
Yarin Gal and Zoubin Ghahramani. 2016.
\newblock Dropout as a bayesian approximation: Representing model uncertainty
  in deep learning.
\newblock In \emph{international conference on machine learning}, pages
  1050--1059. PMLR.

\bibitem[{Guo et~al.(2017)Guo, Pleiss, Sun, and
  Weinberger}]{guo2017calibration}
Chuan Guo, Geoff Pleiss, Yu~Sun, and Kilian~Q Weinberger. 2017.
\newblock On calibration of modern neural networks.
\newblock In \emph{International Conference on Machine Learning}, pages
  1321--1330. PMLR.

\bibitem[{Gururangan et~al.(2020)Gururangan, Marasovi{\'c}, Swayamdipta, Lo,
  Beltagy, Downey, and Smith}]{gururangan-etal-2020-dont}
Suchin Gururangan, Ana Marasovi{\'c}, Swabha Swayamdipta, Kyle Lo, Iz~Beltagy,
  Doug Downey, and Noah~A. Smith. 2020.
\newblock \href {https://doi.org/10.18653/v1/2020.acl-main.740} {Don{'}t stop
  pretraining: Adapt language models to domains and tasks}.
\newblock In \emph{Proceedings of the 58th Annual Meeting of the Association
  for Computational Linguistics}, pages 8342--8360, Online. Association for
  Computational Linguistics.

\bibitem[{Jiang et~al.(2021)Jiang, Araki, Ding, and Neubig}]{jiang2021can}
Zhengbao Jiang, Jun Araki, Haibo Ding, and Graham Neubig. 2021.
\newblock How can we know when language models know? on the calibration of
  language models for question answering.
\newblock \emph{Transactions of the Association for Computational Linguistics},
  9:962--977.

\bibitem[{Kong et~al.(2020)Kong, Jiang, Zhuang, Lyu, Zhao, and
  Zhang}]{kong-etal-2020-calibrated}
Lingkai Kong, Haoming Jiang, Yuchen Zhuang, Jie Lyu, Tuo Zhao, and Chao Zhang.
  2020.
\newblock \href {https://doi.org/10.18653/v1/2020.emnlp-main.102} {Calibrated
  language model fine-tuning for in- and out-of-distribution data}.
\newblock In \emph{Proceedings of the 2020 Conference on Empirical Methods in
  Natural Language Processing (EMNLP)}, pages 1326--1340, Online. Association
  for Computational Linguistics.

\bibitem[{Kuleshov and Liang(2015)}]{kuleshov2015calibrated}
Volodymyr Kuleshov and Percy~S Liang. 2015.
\newblock Calibrated structured prediction.
\newblock \emph{Advances in Neural Information Processing Systems},
  28:3474--3482.

\bibitem[{Kumar et~al.(2018)Kumar, Sarawagi, and Jain}]{kumar2018trainable}
Aviral Kumar, Sunita Sarawagi, and Ujjwal Jain. 2018.
\newblock Trainable calibration measures for neural networks from kernel mean
  embeddings.
\newblock In \emph{International Conference on Machine Learning}, pages
  2805--2814. PMLR.

\bibitem[{Ley et~al.(2021)Ley, Bhatt, and Weller}]{ley2021diverse}
Dan Ley, Umang Bhatt, and Adrian Weller. 2021.
\newblock Diverse, global and amortised counterfactual explanations for
  uncertainty estimates.
\newblock \emph{arXiv preprint arXiv:2112.02646}.

\bibitem[{Li et~al.(2016)Li, Monroe, and Jurafsky}]{li2016understanding}
Jiwei Li, Will Monroe, and Dan Jurafsky. 2016.
\newblock Understanding neural networks through representation erasure.
\newblock \emph{arXiv preprint arXiv:1612.08220}.

\bibitem[{Liu et~al.(2020)Liu, Lin, Padhy, Tran, Bedrax-Weiss, and
  Lakshminarayanan}]{liu2020simple}
Jeremiah~Zhe Liu, Zi~Lin, Shreyas Padhy, Dustin Tran, Tania Bedrax-Weiss, and
  Balaji Lakshminarayanan. 2020.
\newblock Simple and principled uncertainty estimation with deterministic deep
  learning via distance awareness.
\newblock \emph{arXiv preprint arXiv:2006.10108}.

\bibitem[{Liu et~al.(2019)Liu, Ott, Goyal, Du, Joshi, Chen, Levy, Lewis,
  Zettlemoyer, and Stoyanov}]{liu2019roberta}
Yinhan Liu, Myle Ott, Naman Goyal, Jingfei Du, Mandar Joshi, Danqi Chen, Omer
  Levy, Mike Lewis, Luke Zettlemoyer, and Veselin Stoyanov. 2019.
\newblock Roberta: A robustly optimized bert pretraining approach.
\newblock \emph{arXiv preprint arXiv:1907.11692}.

\bibitem[{Lundberg and Lee(2017)}]{lundberg2017unified}
Scott~M Lundberg and Su-In Lee. 2017.
\newblock A unified approach to interpreting model predictions.
\newblock In \emph{Proceedings of the 31st international conference on neural
  information processing systems}, pages 4768--4777.

\bibitem[{Maas et~al.(2011)Maas, Daly, Pham, Huang, Ng, and
  Potts}]{maas2011learning}
Andrew Maas, Raymond~E Daly, Peter~T Pham, Dan Huang, Andrew~Y Ng, and
  Christopher Potts. 2011.
\newblock Learning word vectors for sentiment analysis.
\newblock In \emph{Proceedings of the 49th annual meeting of the association
  for computational linguistics: Human language technologies}, pages 142--150.

\bibitem[{Pereyra et~al.(2017)Pereyra, Tucker, Chorowski, Kaiser, and
  Hinton}]{pereyra2017regularizing}
Gabriel Pereyra, George Tucker, Jan Chorowski, {\L}ukasz Kaiser, and Geoffrey
  Hinton. 2017.
\newblock Regularizing neural networks by penalizing confident output
  distributions.
\newblock \emph{arXiv preprint arXiv:1701.06548}.

\bibitem[{Perez et~al.(2022)Perez, Skalski, Barns-Graham, Wong, and
  Sutton}]{perez2022attribution}
Iker Perez, Piotr Skalski, Alec Barns-Graham, Jason Wong, and David Sutton.
  2022.
\newblock Attribution of predictive uncertainties in classification models.
\newblock In \emph{The 38th Conference on Uncertainty in Artificial
  Intelligence}.

\bibitem[{Ribeiro et~al.(2016)Ribeiro, Singh, and Guestrin}]{ribeiro2016should}
Marco~Tulio Ribeiro, Sameer Singh, and Carlos Guestrin. 2016.
\newblock " why should i trust you?" explaining the predictions of any
  classifier.
\newblock In \emph{Proceedings of the 22nd ACM SIGKDD international conference
  on knowledge discovery and data mining}, pages 1135--1144.

\bibitem[{Schuster et~al.(2019)Schuster, Shah, Yeo, Roberto Filizzola~Ortiz,
  Santus, and Barzilay}]{schuster-etal-2019-towards}
Tal Schuster, Darsh Shah, Yun Jie~Serene Yeo, Daniel Roberto Filizzola~Ortiz,
  Enrico Santus, and Regina Barzilay. 2019.
\newblock \href {https://doi.org/10.18653/v1/D19-1341} {Towards debiasing fact
  verification models}.
\newblock In \emph{Proceedings of the 2019 Conference on Empirical Methods in
  Natural Language Processing and the 9th International Joint Conference on
  Natural Language Processing (EMNLP-IJCNLP)}, pages 3419--3425, Hong Kong,
  China. Association for Computational Linguistics.

\bibitem[{Strumbelj and Kononenko(2010)}]{strumbelj2010efficient}
Erik Strumbelj and Igor Kononenko. 2010.
\newblock An efficient explanation of individual classifications using game
  theory.
\newblock \emph{The Journal of Machine Learning Research}, 11:1--18.

\bibitem[{Sundararajan et~al.(2017)Sundararajan, Taly, and
  Yan}]{sundararajan2017axiomatic}
Mukund Sundararajan, Ankur Taly, and Qiqi Yan. 2017.
\newblock Axiomatic attribution for deep networks.
\newblock In \emph{International Conference on Machine Learning}, pages
  3319--3328. PMLR.

\bibitem[{Wulczyn et~al.(2017)Wulczyn, Thain, and Dixon}]{wulczyn2017ex}
Ellery Wulczyn, Nithum Thain, and Lucas Dixon. 2017.
\newblock Ex machina: Personal attacks seen at scale.
\newblock In \emph{Proceedings of the 26th international conference on world
  wide web}, pages 1391--1399.

\bibitem[{Xu et~al.(2020)Xu, Desai, and
  Durrett}]{xu-etal-2020-understanding-neural}
Jiacheng Xu, Shrey Desai, and Greg Durrett. 2020.
\newblock \href {https://doi.org/10.18653/v1/2020.emnlp-main.508}
  {Understanding neural abstractive summarization models via uncertainty}.
\newblock In \emph{Proceedings of the 2020 Conference on Empirical Methods in
  Natural Language Processing (EMNLP)}, pages 6275--6281, Online. Association
  for Computational Linguistics.

\bibitem[{Yang et~al.(2019)Yang, Dai, Yang, Carbonell, Salakhutdinov, and
  Le}]{yang2019xlnet}
Zhilin Yang, Zihang Dai, Yiming Yang, Jaime Carbonell, Russ~R Salakhutdinov,
  and Quoc~V Le. 2019.
\newblock Xlnet: Generalized autoregressive pretraining for language
  understanding.
\newblock \emph{Advances in neural information processing systems}, 32.

\bibitem[{Zhao et~al.(2021)Zhao, Wallace, Feng, Klein, and
  Singh}]{zhao2021calibrate}
Tony~Z Zhao, Eric Wallace, Shi Feng, Dan Klein, and Sameer Singh. 2021.
\newblock Calibrate before use: Improving few-shot performance of language
  models.
\newblock \emph{arXiv preprint arXiv:2102.09690}.

\end{thebibliography}
